\documentclass[10pt,twocolumn,letterpaper]{article}

\usepackage{wacv}
\usepackage{times}
\usepackage{epsfig}
\usepackage{float}
\usepackage{graphicx}
\usepackage{amsmath}
\DeclareMathOperator*{\argmax}{arg\,max}

\usepackage{multirow}
\usepackage{amssymb}
\usepackage{array}

\usepackage{breqn} 
\usepackage{verbatim}
\graphicspath{{images/}}
\usepackage{subcaption} 

\newcommand{\myline}{\noalign{\hrule height 0.16ex}}

\wacvfinalcopy 

\def\wacvPaperID{286} 

\ifwacvfinal\fi
\begin{document}

\newcommand{\tableimagewidth}{0.12}
\title{Parametric Synthesis of Text on Stylized Backgrounds using PGGANs}

\author{Mayank Gupta\\
Conduent Labs\\
{\tt\small mayank.gupta3@conduent.com}
\and
Abhinav Kumar\\
Conduent Labs\\
{\tt\small abhinav.kumar@conduent.com}
\and
Sriganesh Madhvanath \\
eBay, New York, US\\
{\tt\small srig@acm.org}
}
\maketitle

\ifwacvfinal\thispagestyle{empty}\fi

\begin{abstract}
    We describe a novel method of generating high-resolution real-world images of text where the style and 
    textual content of the images are described parametrically. Our method combines text to image retrieval techniques with progressive 
    growing of Generative Adversarial Networks (PGGANs) to achieve conditional generation of photo-realistic images that reflect specific styles, 
    as well as artifacts seen in real-world images. We demonstrate our method in the context of automotive license plates. We assess the impact of 
    varying the number of training images of each style on the fidelity of the generated style and demonstrate the quality of the generated images using license plate recognition systems. 
    \\ 
    \normalfont{\textbf{Keywords:} Progressive Growing of GANs, Text to Image Synthesis, License Plate Recognition}
\end{abstract}

\section{Introduction \label{sec:Introduction}}
    The synthesis of photo-realistic images of real world objects has several real world applications, such as in graphic design \cite{sbai2018design}, fashion \cite{zhu2017your} and transportation \cite{rodriguez2012data}. In this paper we consider the specific problem of generating images of scene text, where both the `background style' and the textual content of the image may be specified as input parameters. For example, we may want to generate an image of a coffee mug or a T-shirt with a specific background design and text for an e-commerce site, or document images to train or test an OCR system.
    
    Historically this problem has been addressed using computer graphics, in conjunction with manually designed image distortions to simulate real world images. More recently however, this problem has recently been looked into through generative Computer Vision approaches, specifically, Generative Adversarial Networks (GANs). 
   
    GANs are a class of Neural Networks that have been used extensively in Computer Vision to generate low resolution images containing celebrity faces, scenes, birds and flowers, to name a few. More recently, Conditional GANs have been developed for parametric generation of images from other images or from text. However, these applications have almost always been restricted in terms of the resolution of the generated images.

    To achieve parametric generation of highly realistic images with desired background styles and text content, we combine a recently introduced high resolution image generation method known as Progressive Growing of GANs (PGGANs) with our own text encoding technique. We demonstrate our method in the context of automotive license plates. Parametric generation of license plates has important consequences for training of license plate recognition systems. Although we present license plates as an application of our proposed method, the same method would work for generating any kind of stylized text on a background in high-resolution, such as for graphic T-shirts, street signs, and personalized or corporate branded gift items.    
    
    We test the quality of image generation by measuring the performance of OpenALPR, an open source automatic license plate recognition (ALPR) system, as well as a proprietary commercial ALPR system on the generated data. 
    
    The rest of the paper is organized as follows. Section 2 surveys related work in this area. Section 3 describes the proposed method, with its text encoder and the text-to-image PGGAN architecture. Section 4 describes the empirical evaluation of our method, including the datasets and the train-test splits used. Section 5 showcases results and evaluation of real and generated license plate images using ALPR systems. The paper concludes with discussion of results and directions for future work.

\section{Related Work \label{sec:related_work}}
    Several techniques exist in literature for synthetic image generation, which are often used for style transfer and data augmentation. The most obvious method to do so is using rule based image generation techniques \cite{bulan2017segmentation,rodriguez2012data,serrano2013methods}. These methods suffer from many shortcomings. In general, it is not easy to generate photo-realistic images solely using  rule based image generation. Reproducing realistic distortions and noise seen in real world images is something that rule based methods struggle with.
    
    \subsection{Generative Adversarial Networks}
    The most common generative approaches which can synthesize high quality images with realistic distortions are 
    Variational Autoencoders and Generative Adversarial Networks (GANs). 
    Variational Autoencoders \cite{kingma2013auto} use a latent distribution but images produced by these methods are relatively blurry. Sharp images are difficult to obtain by this method. Generative Adversarial Networks (GANs) \cite{goodfellow2014generative} consist of a generator and a discriminator in a game-theoretic setup, where the task of the generator is to generate synthetic images indiscernible from real ones, while the task of the discriminator is to differentiate a generated image from a real one. Hybrid approaches \cite{makhzani2017pixelgan,ulyanov2017takes} have also been proposed in the literature. Unfortunately, all such methods produce images which are relatively less sharp than images produced from GANs \cite{PGGAN}.
    
    \subsection{Conditional GANs}
    Several variations of GANs have been designed recently to function in a conditional manner, on an image-to-image basis \cite{cyclegan} or on a text-to-image basis \cite{reed2016generative}. Reed et al. \cite{reed2016generative} provide a natural language interface to GANs by solving \cite{reed2016learning} a \emph{text-to-image retrieval} problem first. The text input is fed into an LSTM-RNN based network to produce an encoding. A similar encoding is then produced for the image through a CNN. These two are then compared based on a symmetric misclassification loss. The correct pairs of text and image encodings thus come closer together in the common embedding space resulting in text based zero-shot image retrieval. Reed et al. \cite{reed2016learning} use Birds and Flowers datasets and achieves an image retrieval accuracy of around 55-60\%.
    
    A hybrid approach \cite{wang2017adversarial} of conditional license plate image synthesis uses rule based image generation along with CycleGAN \cite{cyclegan}. The authors produce image templates of the required license plate and then add distortions to them using an image-to-image conversion GAN to generate realistic images. 
    
    \subsection{Generation of high resolution images}
    The biggest drawback GANs suffered from was the limited size of images that could be generated in a reasonable amount of training time. However, high-resolution generated images are critical for many applications since a lower resolution reduces the recognizability of the generated image. A license plate from our dataset, for example, is not legible when at $64\times 64$ pixel resolution, but perfectly readable at $256\times 256$. There have been several attempts to generate higher resolution images using GANs.
    Denton et al. \cite{denton2015deep} use a series of conditional GANs to first generate a very low-resolution of the image and then incrementally adds details to the image.
    Zhang et al. \cite{zhang2017stackgan} use stacked GAN where the output of the first GAN and the penultimate layer of the discriminator (as a latent representation) are fed as inputs to the second GAN. 
    
    Karras et al. \cite{PGGAN} achieve high quality $(1024\times 1024)$ image generation by slowly increasing the number of layers in the discriminator and generator. This helps in learning the structure of the dataset gradually from lower resolutions to higher ones and thus the optimum is reached faster and through a more efficient route. Karras et al. \cite{PGGAN} use several novel techniques and achieves a very good result on $1024\times 1024$ image generation. With \cite{PGGAN}, the time taken to train GANs that can produce sharp high-resolution images containing readable characters, has been reduced greatly.

    \subsection{Stabilization of GANs}
    Stabilization of GANs is another issue that has been researched in depth. Gulrajani et al. \cite{WGAN-GP} demonstrate a technique for the stable training using Wasserstein GANs. It introduces a gradient penalty to enforce the Lipschitz constraint on the discriminator. We use this technique for stable training of our own GAN.
    

    \subsection{Loss functions}
    One of the most effective methods for creating well-separated labelled clusters in an embedding space, for a very large number of classes, is the Triplet loss function \cite{Facenet}. This is highly useful for learning latent representations for each class. These latent representations can then be passed to a GAN for conditional generation. The Triplet loss is defined as follows.
    
    Given a function $f$ from datapoints $x_i$'s to an embedding space, the Triplet loss, as defined in Eq. \ref{eq:Trip}, iterates over all $N$ possible triplets $\{a,p,n\}$ in a minibatch, where $a$ is the anchor, $p$, or the positive, is another example of the anchor class and $n$, or the negative, is an example of a different class.
    \begin{equation}\label{eq:Trip}
            L_{apn} = \sum\limits_{i}^{N} \left[||f(x_i^a)-f(x_i^p)||_2^2-||f(x_i^a)-f(x_i^n)||_2^2+\alpha\right]_+
    \end{equation}
    Extensions of the Triplet loss and similar loss functions have been used for multi-modal alignment in a common representation space \cite{reed2016learning,wehrmann2018order}.

    \begin{figure*}[!htb]
            \centering
            \includegraphics[width=0.95\textwidth]{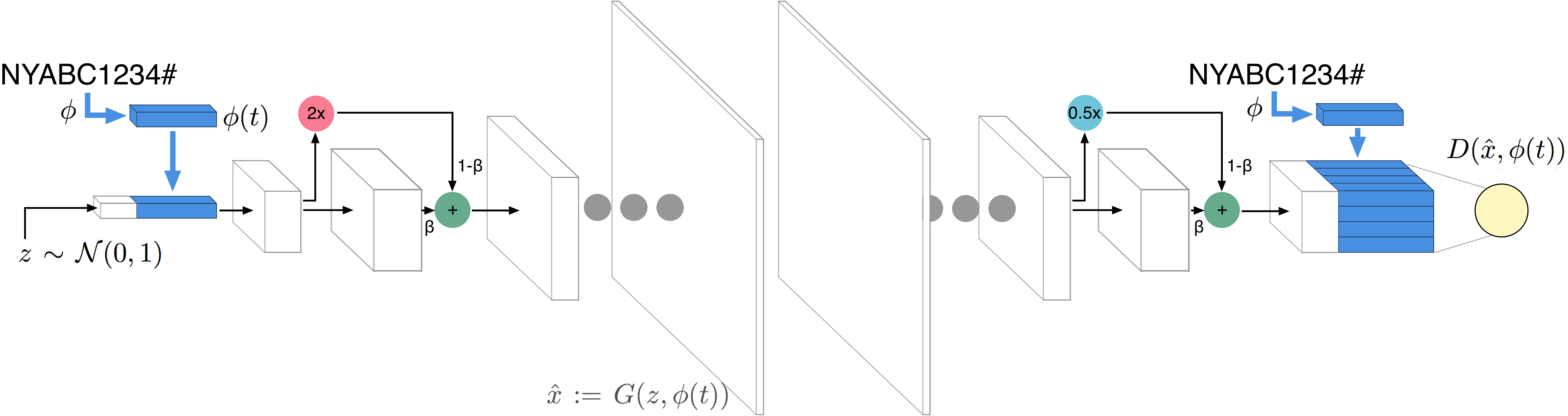}
            \caption{Proposed GAN Architecture. This architecture combines the ideas of progressive growing of GAN \cite{PGGAN} with the text to image interface of \cite{reed2016learning} to get parametric generation of high-resolution images.
            Note that the text encoder $\phi$ is not trained with the GAN framework; it is pre-trained.}
            \label{fig:gan}
    \end{figure*} 
\section{Proposed Solution\label{sec:soln}}

    Since our problem is the parametric generation of images with specific background styles and text content, we assume the input to the system to be a feature vector encoding the style and specifying the content of the output image as fixed length text strings. We assume access to training images featuring the background styles of interest.
    
    License plates of different US states feature different backgrounds. In this context, we simply encode the style as the two letter abbreviation of the state. Thus the New York state license plate number 'ABC 1234' would be simply 'NYABC1234\#'. 
    
    The proposed solution is comprised of two parts, the text encoder and the text-to-image PGGAN.

    \subsection{Text Encoder}
        
        Following the approach of \cite{reed2016learning}, we solve the multi-modal matching problem by mapping the text labels and the images to a common embedding space using two networks. The first one, is a two layered Neural Network with sigmoid activation and a batch-normalization layer and maps text labels to a 512 dimensional common embedding space. The second network \cite{kumar2017neural} is a similar two layered Neural Network which is used jointly with Fisher Vectors of images to map the them to the common embedding space. We use a pair of Triplet losses instead of the misclassification loss as proposed in \cite{reed2016learning} because we found the convergence behavior of Triplet loss to be superior in our experiments.

        For a given minibatch of license plate images $\mathcal{I}$ and corresponding text labels $\mathcal{T}$, using equation \ref{eq:Trip}, we can define the Symmetric Triplet Loss as 
        \begin{equation}\label{eq:symmetric_triplet}
            L_{sym} = L_{itt} + L_{tii}
        \end{equation}
        where, $L_{itt}$ and $L_{tii}$ are two  triplet losses which can be written as
        \begin{equation}\label{eq:I1T2}
            \begin{split}
                L_{tii} &= \sum_{\forall (i_a, t_a) \in (\mathcal{I},\mathcal{T})} \argmax_{A\subset T, |A| = \eta} \phantom{L_{I_1T_2}= \sum_{\forall (i_a, t_a)}} \\
                & \phantom{\sum\sum_{\forall (i_a, t_a)}}\sum_{t_n \in A} \left[d(i_a,t_a) - d(i_a,i_n) + \alpha\right]_+
            \end{split}
        \end{equation}
        and 
        \begin{equation}\label{eq:I2T1}
            \begin{split}
                L_{itt} &= \sum_{\forall (t_a, i_a) \in (\mathcal{T},\mathcal{I})} \argmax_{A\subset I, |A| = \eta} \phantom{L_{I_1T_2}= \sum_{\forall (i_a, t_a)}} \\
                & \phantom{\sum\sum_{\forall (i_a, t_a)}}\sum_{i_n \in A} \left[d(t_a,i_a) - d(t_a,t_n) + \alpha\right]_+
            \end{split}
        \end{equation}
        respectively.
        
        

    \subsection{Network Architecture}
        We modify the network architecture proposed in \cite{PGGAN} for conditional generation of license plates.
               
        The input to the text encoder is given in the form of multi-hot encodings of license plate text label (inclusive of the state), comprising of one-hot encodings for each character. Each license plate number consists of a maximum of 10 characters. For each character, we have 37 choices (26 alphabets, 10 numerals and 1 for absence of any character). Hence, our multi-hot encoding for license plates $t$ is 370 dimensional. This is then passed through the pre-trained text encoder $\phi$ to get a 512 dimensional text encoding $\phi(t)$.
        
        We start building the Generator and Discriminator network from the standard $4 \times 4$ resolution as shown in \cite{PGGAN}. We provide the previously extracted 512-dimensional text encodings $\phi(t)$ as ``label" input to the generator throughout this growing process. In the discriminator network, instead of having 513 outputs, one for real/synthetic prediction and 512 for the labels, we take the penultimate $4 \times 4 \times 512$ layer of features and append a $4 \times 4$ times repeated matrix of 512-dimensional labels $\phi(t)$ along the z-dimension to get a feature map of $4 \times 4 \times 1024$ dimensions \cite{reed2016generative}. We then convolve the final layer over this feature map to output either a $1$ for real images with correct labels or a $0$ for synthetic images and for real images with wrong labels.
        
        We feed the penultimate layer of the discriminator with (a) real images with corresponding text label encodings, (b) generated images with corresponding text label encodings and (c) real images with incorrect text label encodings to improve the association of text and image for the discriminator as described in \cite{reed2016generative}. This is essential for parametric generation of high-resolution images.
        
        Once a sufficient level of training is reached on $4\times 4$ resolution, another layer is gradually ``faded in'' to double the resolution to $8\times 8$ and so on \cite{PGGAN}. The process continues until the GAN is able to generate images of size $256\times256$.
        
        Furthermore, we use WGAN-GP \cite{WGAN-GP} for stabilization. The complete network architecture is shown in figure \ref{fig:gan}.
        
\section{Experimentation \label{sec:expt}}
    In order to evaluate the proposed method, we conducted a series of experiments using US license plate images. 
        \begin{figure}[!htb]
            \centering
            \includegraphics[width=0.4\textwidth]{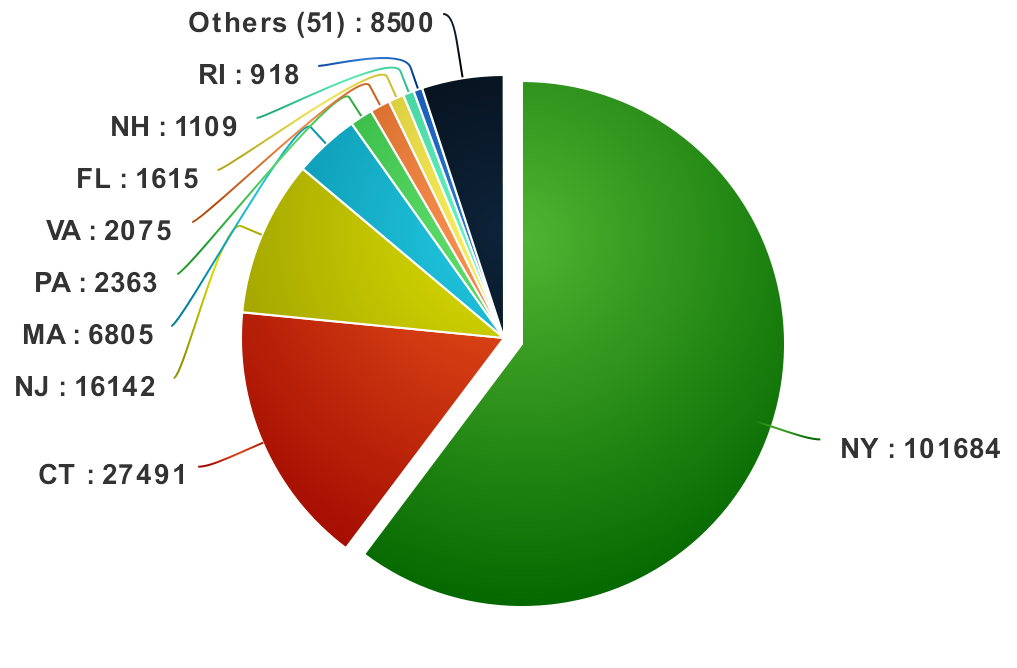}
            \caption{Distribution of 60 styles (states) in license plate database in decreasing order. Note that 3 styles-NY, CT and NJ are dominant in the dataset. (Best viewed in color)}
            \label{fig:pie_chart}
        \end{figure}
    \subsection{Data Splits \label{sec:data_splits}}
        We use a dataset containing about 168,000 grayscale American license plates (70,000 unique plates). The distribution of the data is shown in figure \ref{fig:pie_chart}.

        We then split this randomly into a training set containing 126,000 plates (52,500 unique plates) and a test set containing 42,000 plates (17,500 unique plates). We keep the train and the test set non-overlapping in terms of plate labels.
    
    \subsection{Pre-processing}
        We resize the images to $224\times224$ using bicubic interpolation and zero-padding. Images are resized preserving their aspect ratio in such a way that the larger dimension becomes $224$, so that we do not lose any image content. These images are then centered over and pasted onto an empty $256\times256$ image. 
        
        Our network architecture requires training images of each resolution from $4 \times 4$ up to $256 \times 256$. Therefore, the final $256 \times 256$ images are downsampled to dimensions of powers of $2$ to obtain the lower resolution datasets required.
    
    \subsection{Training}
        We use Tensorflow \cite{abadi2016tensorflow} for all our experiments. We first train the network on 4 Tesla K80 GPUs for 22 days to obtain a `Generic' model for all states. We used the default PGGAN parameters to train.
        
        We then fine-tune this Generic GAN using New York (NY), Virginia (VA) and Missouri (MO) license plates to obtain `Style-tuned' GANs. These states are chosen as there are around a hundred thousand ($O(10^6)$), two thousand ($O(10^3)$) and one hundred and fifty ($O(10^2)$) license plates in our dataset for these states respectively.
    
    \subsection{Evaluation}
        After each GAN's training is complete, the generator model is used to generate synthetic images using the multi-hot encodings of test set license plates. The generated images are then sampled for visual comparison with real plates of the same label. This comparison is a powerful measure of  the generative capabilities of our GAN because the test set is zero shot, and hence has never been seen by either the generator or the discriminator.

\section{Results}\label{sec:results}
    Sample synthetic images generated from different GANs have been shown in table \ref{tab:GANTable}. The noise vector $z$, in figure \ref{fig:gan} is responsible for producing random levels of shadows, contrast levels, skew, blur, as well as different image dimensions. These can be controlled by using techniques like those proposed by Chen et al. \cite{chen2016infogan} 

    \begin{table}[htb]
        \centering
        \begin{tabular}{|>{\centering}m{0.15in}|>{\centering}m{0.75in}|>{\centering}m{0.75in} |>{\centering\arraybackslash}m{0.75in}|}
        \hline
        ~ & Real Image & Generic GAN & Style-tuned GAN \\ \hline
        NY &\includegraphics[width=\tableimagewidth\textwidth]{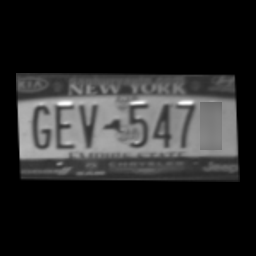}&
        \includegraphics[width=\tableimagewidth\textwidth]{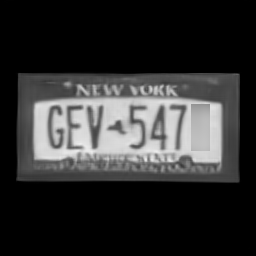}&
        \includegraphics[width=\tableimagewidth\textwidth]{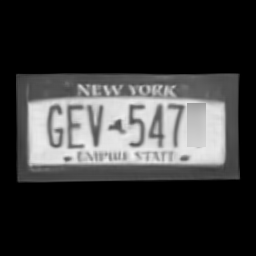}\\ \hline
        VA & \includegraphics[width=\tableimagewidth\textwidth]{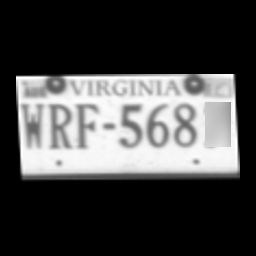}&
        \includegraphics[width=\tableimagewidth\textwidth]{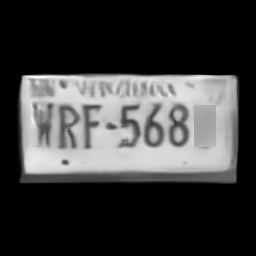}&
        \includegraphics[width=\tableimagewidth\textwidth]{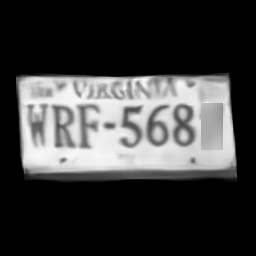}\\ \hline
        MO & \includegraphics[width=\tableimagewidth\textwidth]{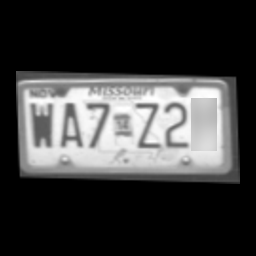} &
        \includegraphics[width=\tableimagewidth\textwidth]{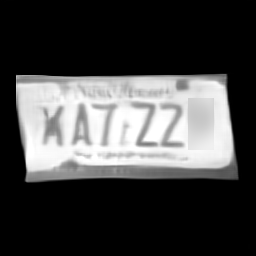} &
        \includegraphics[width=\tableimagewidth\textwidth]{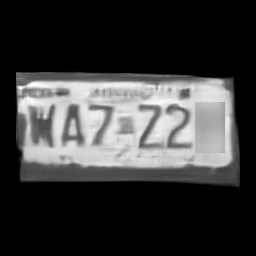}\\ \hline
        \end{tabular}
        \caption{Sample synthetic images generated from Generic and the State Finetuned GANs, and their corresponding real images for NY, VA and MO states. (Some letters have been redacted for privacy)}
        \label{tab:GANTable}
    \end{table}

    \subsection{ALPR Benchmarking}
        One of the ways to quantitatively assess the quality of the generated license plates is by calculating an ALPR system's accuracy on the generated synthetic data and comparing its performance on the real images of the test set. Since these systems are developed specifically to recognize real-world license plate images from different states, their performance on synthetic plates vs real plates allows assessment of the authenticity of both the background style as well as the text. 
        
        We examined the performance of OpenALPR \cite{openalpr}, an open source ALPR system, as well as a proprietary commercial ALPR system on images generated from the Generic GAN as well as the Style-tuned GANs.
    
\begin{figure}[htb]
            \centering
            \includegraphics[width=0.45\textwidth]{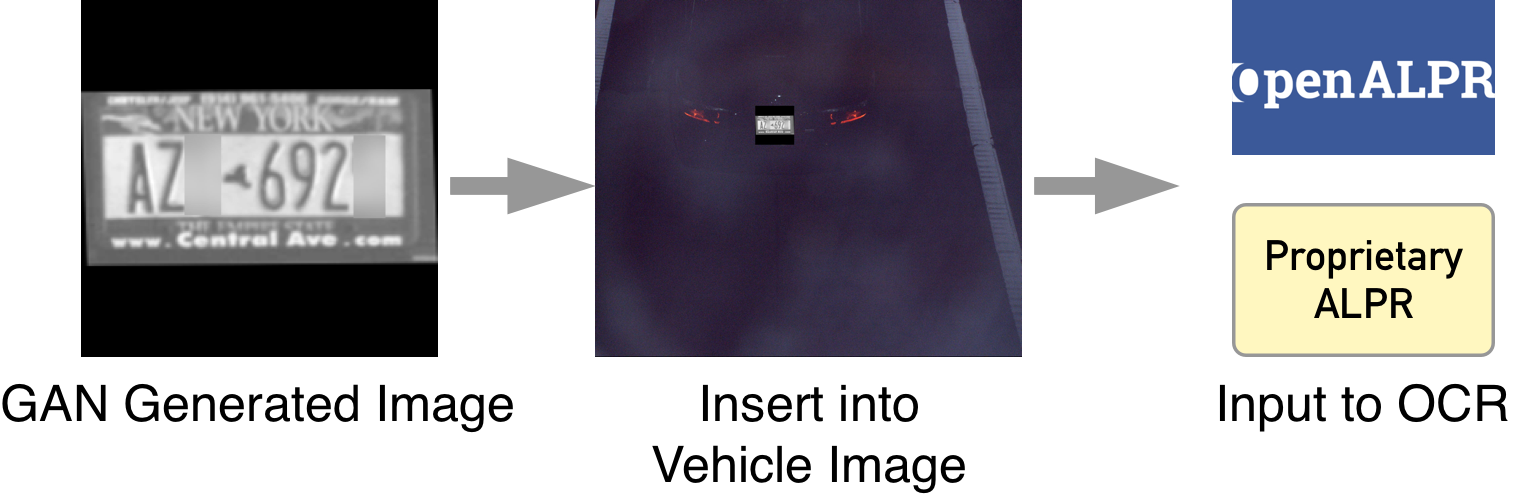}
            \caption{Pasting of the license plate image on the template vehicle image.}
            \label{fig:template_pasted}
        \end{figure}

        Both ALPR systems work only on images containing a full vehicle on the road with a clearly visible license plate. Therefore we pasted all of our test license plate images over the background image of a vehicle to get the required roadside view \cite{rodriguez2012data}. The background vehicle image is kept the same across all the ALPR experiments to ensure a fair comparison. The pasting procedure on the background vehicle template is shown in figure \ref{fig:template_pasted}.
        
        We evaluate performance of the ALPR systems on the following four metrics, wherever applicable, on the real and the synthetic images: (a) \textit{Total Accuracy} - the percentage of complete text labels correctly identified (b) \textit{State Accuracy} - the percentage of states correctly identified, (c) \textit{Word Level Accuracy} - the percentage of license plate numbers correctly identified (d) \textit{Character Level Accuracy} - the percentage of individual licence plate characters correctly identified.

        \subsubsection{OpenALPR}
            OpenALPR \cite{openalpr} returns the license plate number but does not return the state to which the license plate belongs. The performance of OpenALPR is shown in table \ref{tab:ALPR_All}. We do not use the OpenALPR results to draw conclusions as its performance is significantly worse than the commercial ALPR.
            
            \begin{table}[!htb]
                \centering
                \begin{tabular}{|c|c|c|c|}
                \myline
                ~ & \textbf{Images} & Word  & Char \\
                \myline
                All & Real Images & 52.49 & 73.24 \\
                \cline{2-4}
                States & Generic GAN & 38.26 & 68.90 \\
                \myline
                ~ & Real Images &  47.65 & 73.78 \\
                \cline{2-4}
                NY & Generic GAN & 43.19& 73.82\\
                \cline{2-4}
                ~ & NY Finetuned & 46.25 & 74.97\\
                \myline
                ~ & Real Images & 58.66 & 66.45\\
                \cline{2-4}
                VA & Generic GAN & 73.82 & 39.51  \\
                \cline{2-4}
                ~ & VA Finetuned & 45.14& 58.70 \\
                \myline
                ~ & Real Images & 78.94& 84.64\\
                \cline{2-4}
                MO & Generic GAN & 0 & 15.35\\
                \cline{2-4}
                ~ & MO Finetuned & 10.53 & 41.67 \\
                \myline
                \end{tabular}
                \caption{\label{tab:ALPR_All}Accuracies (in \%) obtained after running OpenALPR on real images as well as synthetic images.}
            \end{table} 
    
        \subsubsection{Proprietary ALPR}
            The proprietary ALPR system returns the state along with the plate number. It thus evaluates the extent of style transfer more effectively as compared to OpenALPR. The results on the original images and synthetic images generated through Generic GAN are shown in table \ref{tab:XLPR_All}. The table also shows the performance of the proprietary ALPR on plates generated from Style-tuned GANs.
            
            \begin{table}[!htb]
                \centering
                \begin{tabular}{|c|c|c|c|c|c|}
                \myline
                ~ & \textbf{Images} & Total  & State  & Word  & Char \\
                \myline
                All & Real Images & 85.96 & 88.46 & 90.64 & 94.40 \\
                \cline{2-6}
                States & Generic GAN & 67.84 & 85.23 & 68.66 & 88.96 \\
                \myline
                ~ & Real Images & 92.38 & 95.20 & 93.61 & 95.94 \\
                \cline{2-6}
                NY & Generic GAN & 84.20 & 94.50 & 84.19 & 93.83 \\
                \cline{2-6}
                ~ & NY Finetuned & 89.08 & \textbf{95.96} & 89.08 & 95.38 \\
                \myline
                ~ & Real Images & 78.59 & 81.62 & 88.44 & 92.28  \\
                \cline{2-6}
                VA & Generic GAN & 32.19 & 69.71 & 35.23 & 81.47  \\
                \cline{2-6}
                ~ & VA Finetuned & 75.62 & 81.14 & 76.95 & 88.68  \\
                \myline
                ~ & Real Images & 94.73 & 100 & 94.73 & 96.92  \\
                \cline{2-6}
                MO & Generic GAN & 0 & 0 & 0 & 25.43 \\
                \cline{2-6}
                ~ & MO Finetuned & 13.63 & 52.27 & 13.63 & 57.19 \\
                \myline
                \end{tabular}
                \caption{\label{tab:XLPR_All}Accuracies (in \%) obtained after running the Proprietary ALPR on real as well as synthetic images.}
            \end{table}
            
            Looking at the State Accuracies in table \ref{tab:XLPR_All}, the Generic GAN is quite successful at fooling the ALPR system on the state classification task. It is not as successful with the license plate numbers.
            
\begin{figure*}[t]
                \centering
                \subcaptionbox{Total Accuracy Plot}[.48\linewidth][c]{%
                    \includegraphics[width=.47\linewidth,height=.22\linewidth]{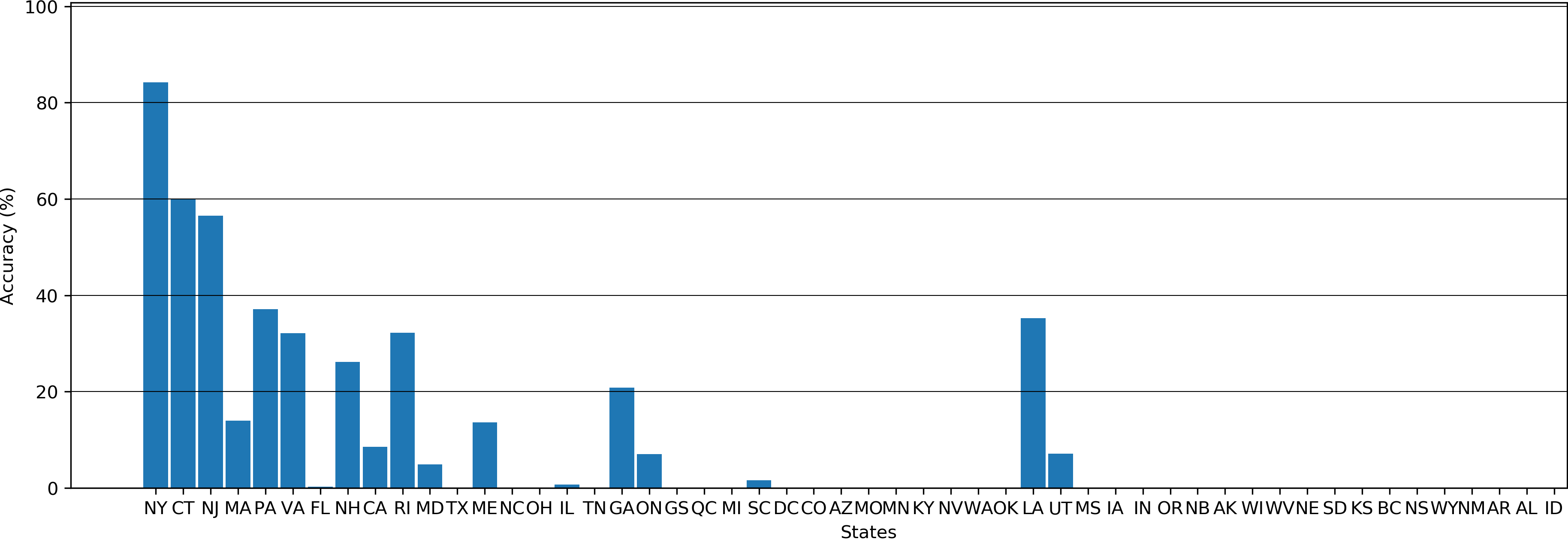}
                    }\;
                \subcaptionbox{State Accuracy Plot}[.48\linewidth][c]{%
                    \includegraphics[width=.47\linewidth,height=.22\linewidth]{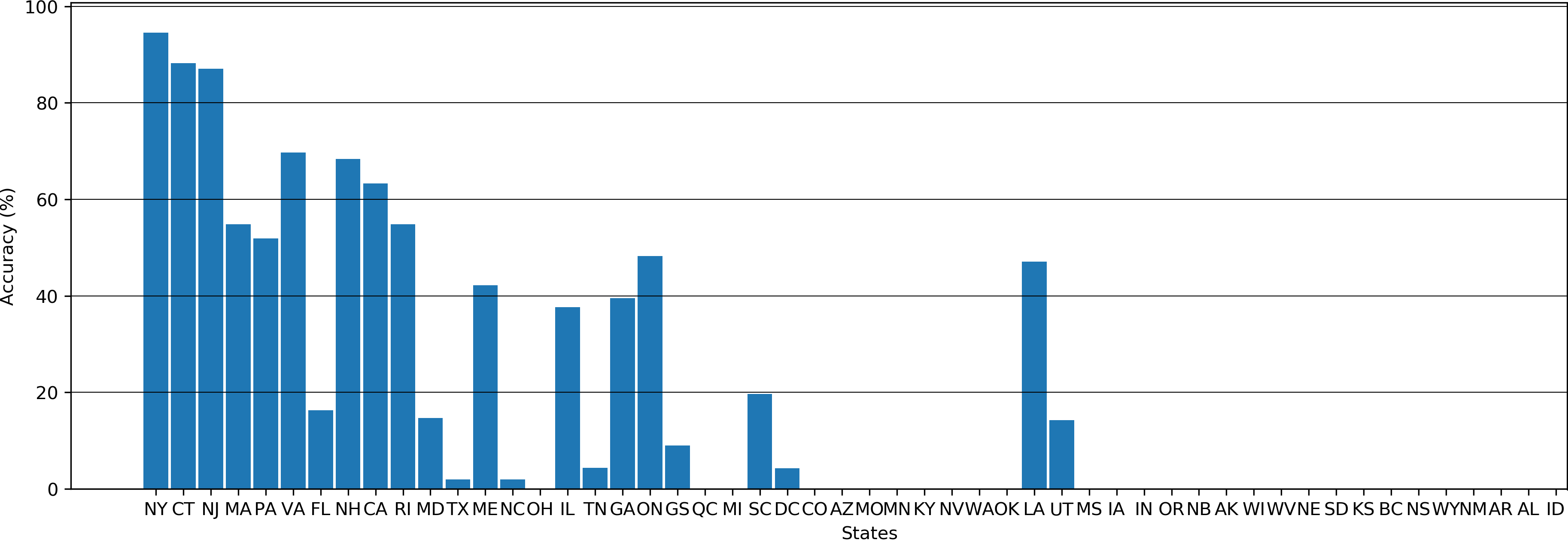}
                    }
                \bigskip
                
                \subcaptionbox{Word Accuracy Plot}[.48\linewidth][c]{%
                    \includegraphics[width=.47\linewidth,height=.22\linewidth]{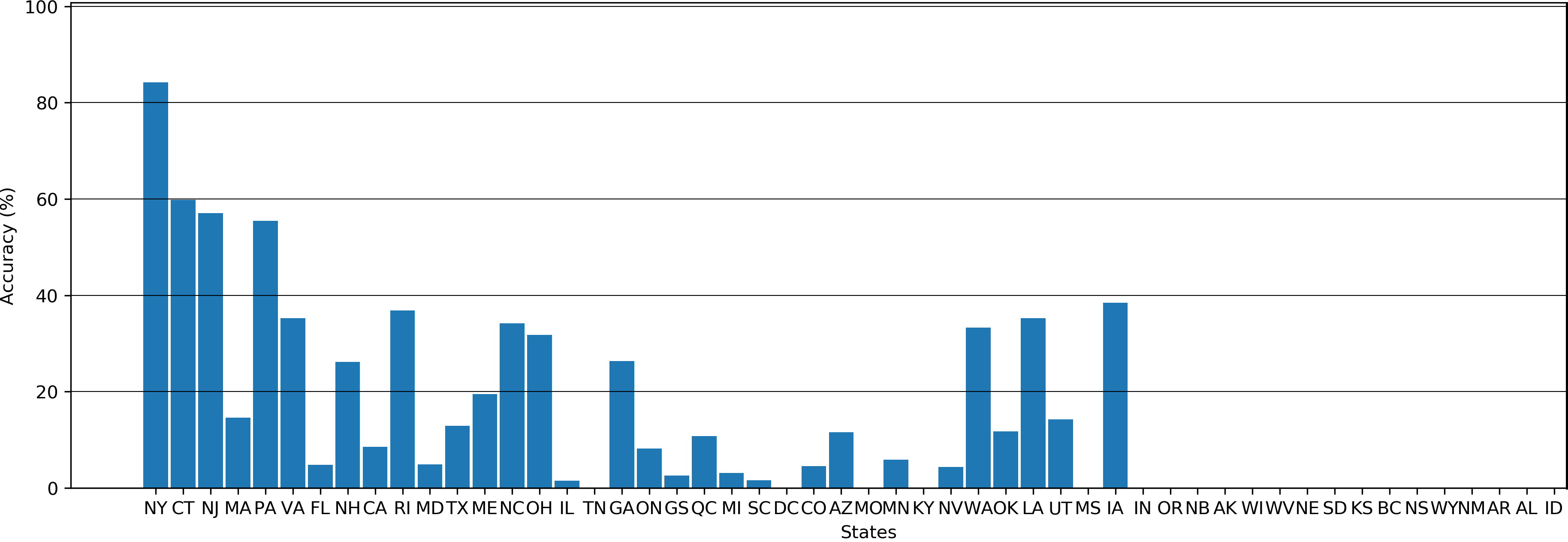}}\;
                \subcaptionbox{Char Accuracy Plot}[.48\linewidth][c]{%
                    \includegraphics[width=.47\linewidth,height=.22\linewidth]{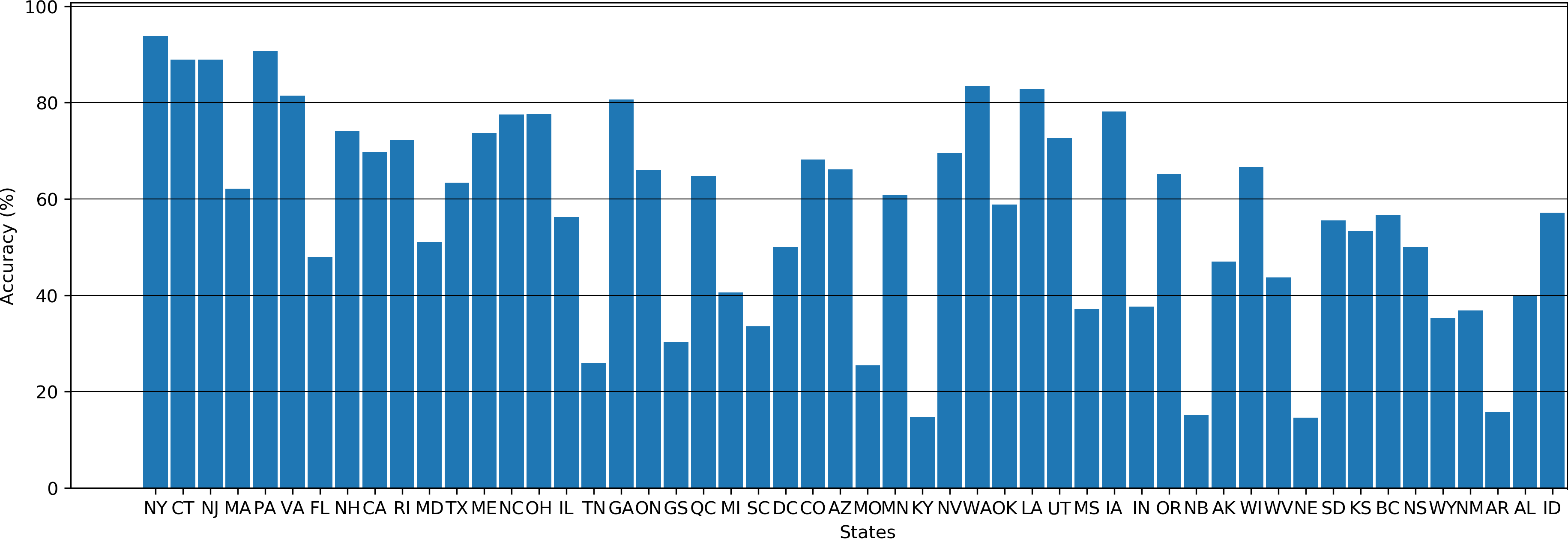}}
                \caption{\label{fig:XLPR_States}Proprietary ALPR's state-wise performance on images generated by the Generic GAN model. States are on the horizontal axis and are sorted in descending order by number of datapoints in test set.}
            \end{figure*}

            Figure \ref{fig:XLPR_States} shows accuracy metrics for each state in descending order of the number of datapoints for that state in the test set. The plot suggests a correlation between the accuracies obtained and the amount of data shown to the GAN during training for each state. Also, data-insufficiency affects state accuracies more as compared to character accuracy. The fact that the character accuracy is not affected much can be explained by the ease of generating digits regardless of desired style. Even if the GAN outputs the desired character in a different font, it would still be read correctly.
            
            As seen in table \ref{tab:XLPR_All}, finetuning the Generic GAN model on different states leads to increase in total, state, word as well as character accuracies on test images of that state. Also, the VA-tuned GAN trained on $O(10^3)$ datapoints has a very similar overall performance to the NY-tuned GAN which has been trained on $O(10^6)$ datapoints. However, there is a sharp dip in performance with the MO-tuned GAN which was finetuned on $O(10^2)$ datapoints. This suggests that for this problem, fine tuning our GAN network separately for each style using $O(10^3)$ samples is helpful to produce much more realistic images for that style.
                                    
\begin{figure}[htb]
                \centering
                \subcaptionbox{Original}[.3\linewidth][c]{     \includegraphics[width=.28\linewidth]{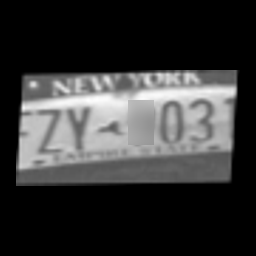}
                    }\;
                \subcaptionbox{Generic GAN}[.3\linewidth][c]{ \includegraphics[width=.28\linewidth]{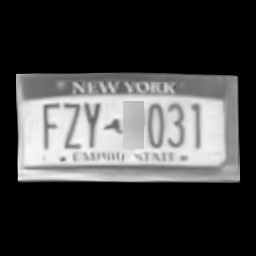}
                    }\;
                \subcaptionbox{NY Finetuned GAN}[.3\linewidth][c]{%
                    \includegraphics[width=.28\linewidth]{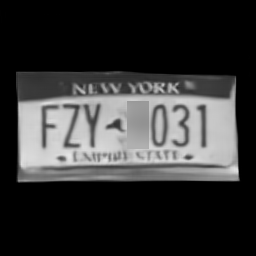}}
                \caption{\label{fig:NY_plates}Correctional behaviour of GAN.}
        \end{figure}

        It is interesting to note in table \ref{tab:XLPR_All} is that the proprietary ALPR does better at state recognition with the generated images (from the NY-tuned GAN) compared to real images. This is because the GAN corrects any ROI segmentation errors and capture defects that occasionally occur in real-world images. This  behavior of the GAN is seen in figure \ref{fig:NY_plates} where the wrongly cropped real image is generated correctly. This may also be seen in the first row of table \ref{tab:GANTable} where the gloss and noise around ``NY GEV547\_" has been removed in the generated plates.

    \subsection{SSIM Benchmarking}
        We also evaluate the quality of the generated images by comparing them to their real counterparts using the Structural Similarity Index Metric (SSIM) \cite{wang2004image}. SSIM returns a number between -1 and 1 for similarity measure. The closer the output is to one, the more similar the two images are. We compared all fake images against their real counter parts. The results of this comparison are shown in table \ref{tab:ssim}. While popular as a image similarity measure, SSIM results do not provide much information about the style transfer and thus are not as useful as the ALPR results. They are included here for completeness.
        
        We do not use the RMSE measure for this comparison as this metric fails to look past realistic distortions generated by the GAN. Moreover, in case the real image itself is highly noisy, skewed or wrongly segmented and the synthetic image is of good quality, RMSE measure reports a high error which is undesirable.
  
        \begin{table}[!htb]
            \centering
            \begin{tabular}{|c|c|c|c|c|}
            \myline
            ~ & All  & NY & VA & MO\\
            ~ & Images & Images & Images & Images\\
            \myline
            Real Images & 0.838 & 0.819 & 0.852 & 0.857\\
            \myline
            Generic GAN & 0.712 & 0.709 & 0.730 & 0.680\\
            \myline
            NY Finetuned & - & 0.707 & - & -\\
            \myline
            VA Finetuned & - & - & 0.733 & -\\
            \myline
            MO Finetuned & - & - & - & 0.725 \\
            \myline
            \end{tabular}
            \caption{\label{tab:ssim}SSIM results of synthetic images taking real images as full-reference. If there are multiple real images of the same label, one of them is chosen at random.}
        \end{table}

    \subsection{Limitations}
        There are certain limitations to the proposed GAN training procedure. Unsurprisingly, the main limitation is that the performance of the model is poor on styles with too few data points. This is evident from the Generic GAN model's performance in figure \ref{fig:XLPR_States}.
        
        The model also does poorly on non-standard  formats of license plates such as ``NY 1234H" or ``NY KATE" or ``CT ROCKER".
        
        \begin{figure}[!htb]
            \centering
            \subcaptionbox{NY ABC1234}[.28\linewidth][c]{%
                \includegraphics[width=.28\linewidth]{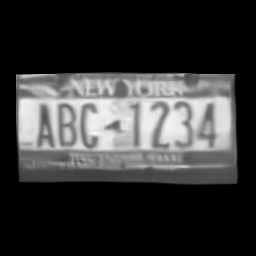}
                }\;
            \subcaptionbox{NY A12BC34}[.28\linewidth][c]{%
                \includegraphics[width=.28\linewidth]{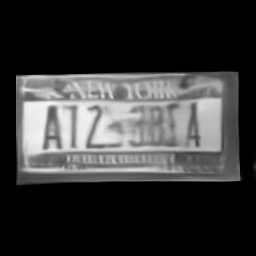}
                }\;
            \subcaptionbox{NY 00QD02}[.28\linewidth][c]{%
                \includegraphics[width=.28\linewidth]{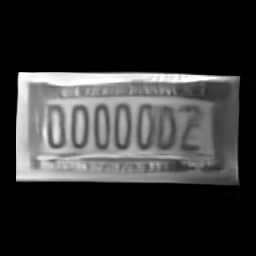}
                }\;
                
            \subcaptionbox{NY 0M11E}[.28\linewidth][c]{%
                \includegraphics[width=.28\linewidth]{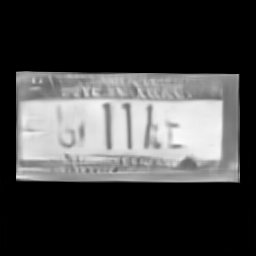}
                }\;
            \subcaptionbox{AZ AE94\_26}[.28\linewidth][c]{%
                \includegraphics[width=.28\linewidth]{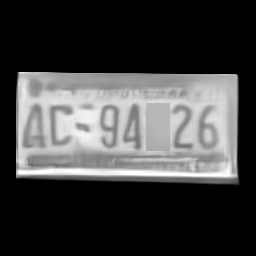}
                }
            \subcaptionbox{AZ SOLACE\_ }[.28\linewidth][c]{%
                \includegraphics[width=.28\linewidth]{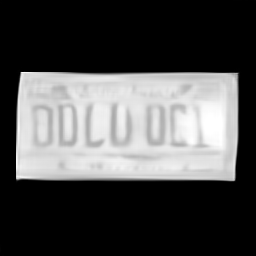}}
            \caption{\label{fig:Limitations}Limitations of our Generic GAN model.} 
        \end{figure}
        
        
        The word ``ABC1234" in the plate, in figure \ref{fig:Limitations}(a), ``NY ABC1234" is compliant with one of the prevalent word formats in NY style and therefore, it is generated perfectly. However, ``NY A12BC34" in figure \ref{fig:Limitations}(b), is not as good because it is not in one of the frequently seen word formats. However, it is still better than (d), as it is closer to the common word formats than (d) is. Images in figures \ref{fig:Limitations}(c) and  \ref{fig:Limitations}(d) are of extremely rare word formats in NY style and therefore have not been generated properly.
        
        The plate in figure \ref{fig:Limitations}(e) belongs to the standard Arizona (AZ) style, a state with only 150 samples in the entire dataset. In the generated plate (e), we see that most of the characters have been generated correctly, however, the style transfer is lacking. The image in figure \ref{fig:Limitations}(f) corresponds to very rare word format and in addition, happens to be from a state with very little data. Hence, both style and textual content are poorly synthesized.
        
        We may conclude that GANs require sufficient data in order to learn a style well. In addition, text whose formats are rarer, are much more difficult for GANs to generate.

\section{Conclusions}
    In this paper, we described a novel method for parametric generation of high resolution license images with specific background styles and text content. Our method combines a text encoding scheme for styles and target text trained using a Symmetric Triplet loss function, with PGGANs for high fidelity, high resolution image synthesis. We described the application of our method to the synthesis of license plate images from different US states, and evaluated the synthesized images using ALPR systems as well as visual inspection and SSIM measures. 
    
    We showed that given sufficient samples of each style, a style-tuned GAN outperforms a generic GAN that is trained on all styles, and produces high quality images that a commercial ALPR system treats on par with real images. In fact, the GAN corrects issues of poor ROI segmentation and other artifacts found in real world images. However the generic GAN still effects reasonable style transfer for the ALPR to recognize the state. 
    
    We also examined some of the limitations of the approach. Not surprisingly, the quality of synthesized images suffers when there are too few samples of a given style in the training data. It is also interesting that rare label patterns are generated much worse than common ones, because the nature of the dataset is such that the trained GAN is not accustomed to seeing certain characters in certain positions for a given states.
    
    Our current method uses a single GAN to generate images that capture both the intended style and text. A potential future direction is to explore cascades of GANs that effect style transfer and text synthesis in separate stages, allowing us to (for example) use different datasets for each task. Going beyond text on background styles to more complex scene text synthesis is another interesting direction.



    \section*{Acknowledgements \label{sec:ack}}
        The authors would like to thank Sriranjani Ramakrishnan and Aishwarya Gupta for their valuable inputs and fruitful discussions. Ivan Novotony and Giri Srinivasan were instrumental in sorting the infrastructural issues.

{\small
\bibliographystyle{ieee}
\bibliography{biblio}
}
\onecolumn
\end{document}